\documentclass[letterpaper]{article} 
\usepackage{aaai2026}  
\usepackage{times}  
\usepackage{helvet}  
\usepackage{courier}  
\usepackage[hyphens]{url}  
\usepackage{graphicx} 
\usepackage{natbib}  
\usepackage{caption}
\usepackage{amsmath,amssymb}
\usepackage{soul,color}
\usepackage{lineno} 
\usepackage{subcaption}
\usepackage[ruled,vlined]{algorithm2e}
\usepackage{booktabs} 
\usepackage{alltt} 
\usepackage{tikz}
\usepackage{xcolor} 
\usepackage{enumitem}

\definecolor{barblue}{RGB}{31,119,180}
\definecolor{barorange}{RGB}{255,127,14}

\makeatletter\def\@listi{\leftmargin\leftmargini \topsep .5em \parsep .5em \itemsep .5em}
\def\@listii{\leftmargin\leftmarginii \labelwidth\leftmarginii \advance\labelwidth-\labelsep \topsep .4em \parsep .4em \itemsep .4em}
\def\@listiii{\leftmargin\leftmarginiii \labelwidth\leftmarginiii \advance\labelwidth-\labelsep \topsep .4em \parsep .4em \itemsep .4em}\makeatother

\newcounter{checksubsection}
\newcounter{checkitem}[checksubsection]

\newcommand{\checksubsection}[1]{%
  \refstepcounter{checksubsection}%
  \paragraph{\arabic{checksubsection}. #1}%
  \setcounter{checkitem}{0}%
}

\newcommand{\checkitem}{%
  \refstepcounter{checkitem}%
  \item[\arabic{checksubsection}.\arabic{checkitem}.]%
}
\newcommand{\question}[2]{\normalcolor\checkitem #1 #2 \color{blue}}
\newcommand{\ifyespoints}[1]{\makebox[0pt][l]{\hspace{-15pt}\normalcolor #1}}

\usepackage{newfloat}
\usepackage{listings}
\DeclareCaptionStyle{ruled}{labelfont=normalfont,labelsep=colon,strut=off} 
\title{HCFSLN: Adaptive Hyperbolic Few-Shot Learning for Multimodal Anxiety Detection}

\author {
    Aditya Sneh\textsuperscript{\rm }\thanks{Corresponding author: adityas19@iiserb.ac.in},
    Nilesh Kumar Sahu\textsuperscript{\rm },
    Anushka Sanjay Shelke\textsuperscript{\rm },
    Arya Adyasha\textsuperscript{\rm },
    Haroon R. Lone\textsuperscript{\rm }
}
\affiliations {
    \textsuperscript{\rm }Indian Institute of Science Education and Research Bhopal\\
       \{adityas19, nilesh21, anushka19, arya20, haroon\}@iiserb.ac.in
}

\usepackage{bibentry}
\usepackage{amssymb}
\usepackage{adjustbox}
\usepackage{array}

\begin{document}

\maketitle

\begin{abstract}
Anxiety disorders impact millions globally, yet traditional diagnosis relies on clinical interviews, while machine learning models struggle with overfitting due to limited data. Large-scale data collection remains costly \& time-consuming, restricting accessibility. To address this, we introduce Hyperbolic Curvature Few-Shot Learning Network (HCFSLN)—a novel Few-Shot Learning (FSL) framework for multimodal anxiety detection, integrating speech, physiological signals, and video data. HCFSLN enhances feature separability through hyperbolic embeddings, cross-modal attention, and an adaptive gating network, enabling robust classification with minimal data. We collected a multimodal anxiety dataset from 108 participants and benchmarked HCFSLN against six FSL baselines, achieving 88\% accuracy outperforming best baseline by 14\%. These results highlight the effectiveness of hyperbolic space for modeling anxiety-related speech patterns and demonstrate FSL’s potential for anxiety classification. 

\end{abstract}

\section{Introduction}

Anxiety disorder is one of the most prevalent mental health conditions, affecting over 301 million people worldwide \cite{javaid2023epidemiology}. Despite its widespread prevalence and substantial impact, detection still primarily relies on clinical diagnosis \cite{vismara2020peripheral}, which often requires multiple visits and can be influenced by variability in clinical judgment and diagnostic criteria.

Recent studies have investigated the use of wearable sensors for detecting and predicting mental health conditions by monitoring physiological signals \cite{sahu2024wearable}. These sensors capture physiological data, which, when combined with self-reported measures, serve as inputs for machine learning and deep learning models. In these studies, participants engage in controlled anxious activities while physiological signals, such as heart rate variability and skin conductance, are recorded. Beyond physiological signals, researchers have also integrated multimodal data sources, including audio and video, to enhance mental health predictions \cite{yoon2022d}. However, many of these studies are constrained by small or imbalanced datasets, which increases the risk of overfitting~\cite{islam2024comprehensive}. Furthermore, publicly available multimodal datasets for mental health research remain scarce, hindering reproducibility and large-scale advancements in the field.

\begin{figure}[t]
    \centering
\includegraphics[scale=0.5]{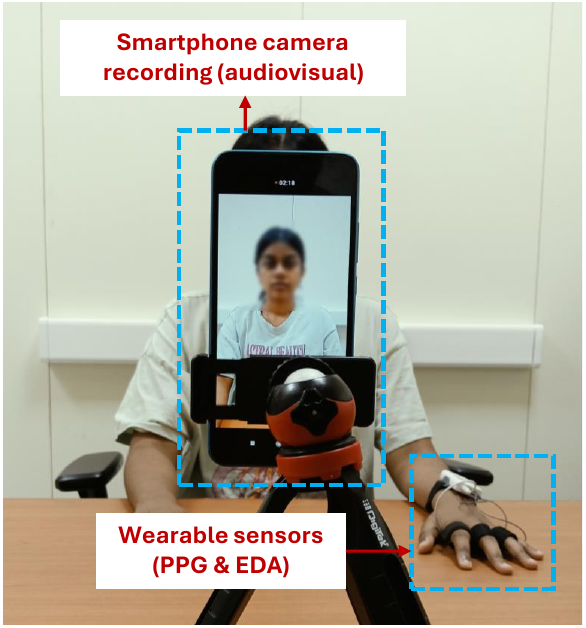}
    \caption{Experimental setup. }
    \vspace{-2em}
    \label{fig:study}
\end{figure}

Existing studies on machine learning model performance suggest that training robust mental health prediction models requires at least 1,000 participants and 100,000 data points to ensure effective learning and evaluation \cite{zantvoort2024estimation}. However, collecting such large datasets is both time-consuming and financially burdensome. While data augmentation techniques are commonly used to compensate for small datasets, their application to physiological signals and multimodal data (such as audio and video) can introduce artifacts, distort underlying patterns, and ultimately degrade model performance \cite{el2024physiological}. To address the challenge of small training samples in anxiety classification, we explore Few-Shot Learning (FSL), a technique that enables models to achieve high accuracy with minimal data \cite{feng2021few}.


To demonstrate the usability of FSL and address the lack of publicly available datasets combining physiological and audiovisual data, we conducted a study in an Asian country to highlight FSL’s relevance in anxiety research. We collected data from 108 participants, who performed a speech activity in front of a smartphone camera (see Figure \ref{fig:study}). The dataset includes physiological signals such as Photoplethysmography (PPG) and Electrodermal Activity (EDA) from wearable sensors along with audiovisual recordings. Additionally, self-reported assessments were collected to classify participants as anxious or non-anxious.

Moreover, this paper presents, Hyperbolic Curvature Few-Shot Learning Network (HCFSLN), a novel FSL framework with Hyperbolic Curvature for anxiety detection. 
HCFSLN enhances feature separability through hyperbolic embeddings, cross-modal attention, and an adaptive gating network, enabling robust classification with minimal data. It outperforms six state-of-the-art (SOTA) FSL baseline models on two different datasets, achieving 88\% accuracy in the 1-shot setting with audio modality.  These results highlight the advantage of hyperbolic space in modeling anxiety-related speech patterns. Our key contributions are:
\begin{itemize}[itemsep=0em,parsep=0em]
\item We propose \textbf{HCFSLN} and systematically benchmark recent few-shot learning models using Euclidean, hyperbolic, and hyperspherical embeddings, and demonstrate that our proposed hyperbolic-based framework consistently outperforms these baselines. To foster reproducibility, we release our code and dataset.
\item We present the \textbf{Multi-Modal Anxiety Dataset (M2AD)}, a real-world dataset comprising physiological (PPG, EDA) and audiovisual signals from 108 participants in a low-middle-income country. This resource addresses a key gap in anxiety research and is made publicly available for further study. 

\end{itemize}

\section{Related Work}

Studies on anxiety detection have explored various modalities, like EDA, Electrocardiogram (ECG), audio, video, etc., to capture behavioral and physiological cues \cite{mo2022sff}. Electroencephalogram (EEG) and ECG provide detailed neural and cardiac insights but require costly equipment and controlled conditions, limiting scalability. However, wearable sensors like PPG and EDA offer a low-cost, unobtrusive solution for continuous physiological monitoring. Moreover, modalities like audio and video from ubiquitous audiovisual cameras present behavioral features.

Various deep-learning algorithms have been explored for anxiety detection. For example, EEG-based approaches use models like EmotionNet \cite{daadaa2024emotionnet}, a CNN-LSTM hybrid to capture anxiety-related brain activity. ECG based methods analyze heart rate variability using techniques such as probabilistic binary pattern  extraction \cite{baygin2024automated} and tunable q-factor wavelet transforms \cite{jha2020cardiac}. Similarly, \citet{mo2024multimodal} proposed MMD-AS, which employs an Improved Fireworks Algorithm for feature selection in smartphone-based anxiety detection. Likewise, \citet{toto2021audibert} proposed speech-based models like AudiBERT, which incorporate acoustic features for voice-based mental health detection. However, to train these deep learning models, we need a large annotated dataset.

To address the limitations of small datasets in the mental health domain, FSL has emerged as a promising approach in mental health assessment, enabling classification with minimal labeled data. Recent studies have employed FSL techniques to detect stress and anxiety conditions. For example, a Siamese Neural Network with Wasserstein distance minimization has been applied to stress classification \cite{feng2022few}, effectively handling distribution mismatch across individuals. Similarly, the Spatiotemporal Feature Fusion for Detecting Anxiety framework, which integrates 3D-CNN + LSTM for feature extraction, enhances multimodal anxiety detection by leveraging few-shot learning for improved generalization \cite{mo2022sff}. Additionally, a meta-learning-based stress category detection framework use an encoder module with a Dependency Graph Convolutional Network, an induction module using a Mixture of Experts' mechanisms, and a relation module for similarity assessment, demonstrating strong generalization in low data scenarios \cite{wang2022meta}. 

Although FSL research has explored various embedding spaces—including Euclidean \cite{li2023euclidean}, hyperbolic \cite{moreira2024hyperbolic}, and hyperspherical \cite{ding2022few}—there remains a significant gap in understanding which geometry works best for physiological‑only, audiovisual‑only, or multimodal anxiety data across different shot scenarios. A study show that hyperbolic embeddings do not always outperform Euclidean ones; performance often depends heavily on training configuration and embedding radius \cite{moreira2024hyperbolic}. While surveys emphasize hyperbolic geometry’s strength in representing hierarchical or low‑dimensional structures and its robustness in low‑sample settings~\cite{mettes2024hyperbolic}, they offer limited guidance on modality‑specific embedding choices. Consequently, no prior work clarifies which embedding space works best for physiological‑only, audiovisual‑only, or multimodal anxiety data, especially under different shot scenarios. Our study fills this gap via a comprehensive comparison across embeddings, modality sets, and shot configurations.

\begin{figure*}[!t]
    \centering
    \begin{center}
        \includegraphics[width=1\linewidth]{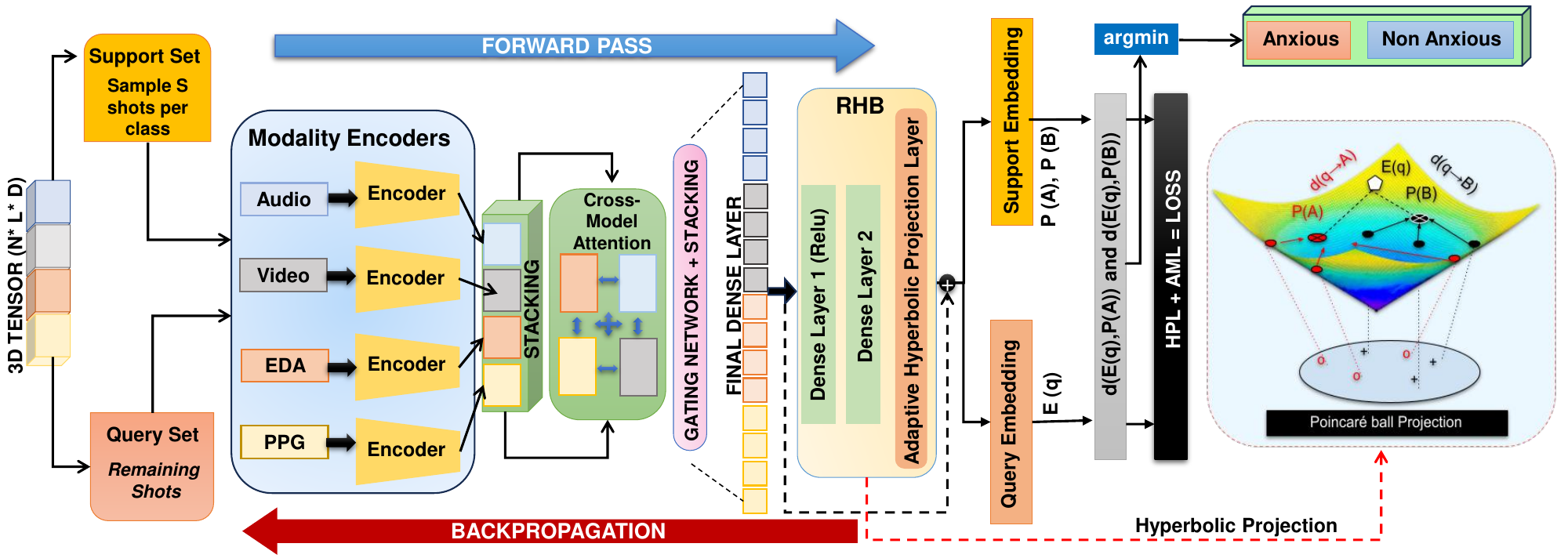}
    \end{center}
    \caption{Architecture of the proposed HCFSLN framework.}
    \label{fig:master-diagram}
\end{figure*}

\section{Proposed Method}

This section discusses HCFSLN (see Figure~\ref{fig:master-diagram}), a hyperbolic curvature-aware framework for multimodal FSL, leveraging hyperbolic embeddings, cross-modal attention, and adaptive gating to improve feature fusion and class separability.

\subsection{Preliminaries}

FSL classifies new samples with minimal labeled data using \textbf{episodic training}, where a model learns from a sequence of binary (i.e., anxious or non-anxious) classification tasks. Each episode consists of a support set
The support set is defined as $\mathcal{S} = \{ (\mathbf{x}_j, y_j) \}_{j=1}^{N_S}$ with $N_S = 2K$ labeled examples (i.e., $K$ examples per class), and the query set is $\mathcal{Q} = \{ (\mathbf{x}_k, y_k) \}_{k=1}^{N_Q}$ with \(N_Q = 2B\) unlabeled examples (i.e., \(B\) examples per class). Here, \(\mathbf{x}_j \in \mathbb{R}^d\) is a \(d\)-dimensional multimodal feature vector, and \(y_j \in \{0,1\}\) is its binary class label. Similarly, \(\mathbf{x}_k\) is a \(d\)-dimensional query feature vector with its corresponding label \(y_k\). During training, $y_k$ are available for loss computation, but during inference, FSL models predict query labels without supervision. Each data sample \(\mathbf{x}\) consists of features from \(M\) different modalities. It is represented as
\begin{equation} \label{eq:sample_modality}
\mathbf{x} = \big[ \mathbf{x}^{(1)}, \mathbf{x}^{(2)}, \dots, \mathbf{x}^{(M)} \big]
\end{equation}

where \(\mathbf{x}^{(m)}\) is the feature vector from the \(m\)-th modality, such as audio, PPG, EDA, or video. The challenge lies in integrating heterogeneous modalities while preserving discriminative representations in small dataset settings and mitigating modality-specific noise and cross-modal interference. To address this, we learn an embedding function $f_\theta: \mathbb{R}^d \to \mathbb{B}^d$ that maps features from Euclidean space \(\mathbb{R}^d\) to a hyperbolic space \(\mathbb{B}^d\) (Poincaré ball model) to promote compact intra-class clustering and inter-class separation. A \textbf{modality cross-attention mechanism} with a \textbf{gating network} (filters noise) mitigates irrelevant noise during training \cite{jia2023multi}. Unlike Euclidean-based FSL models, our approach employs an \textbf{adaptive curvature parameter} (controls hyperbolic geometry) denoted by \(\alpha > 0\) to capture complex non-Euclidean relationships. The objective is to develop a robust few-shot framework that integrates heterogeneous multimodal features in hyperbolic space, leveraging adaptive curvature and prototype-based classification for enhanced clustering and separation.

\subsection{Multimodal Feature Fusion}
Before any hyperbolic processing or prototype computation, the model obtains a unified multimodal representation \(\mathbf{h}\) that effectively integrates all modalities.

\subsubsection{Cross-Modal Attention} Each modality \(\mathbf{x}^{(m)}\) undergoes feature extraction using dedicated \textbf{modality encoders}, which capture meaningful representations from raw multimodal inputs. Given an input sample with multiple modalities, the initial feature vector shown in Eq~\eqref{eq:sample_modality}, where each $\mathbf{x}^{(m)}$ is processed through a modality-specific encoder $f_m(\cdot)$, which includes two convolutional layers (Conv1 with kernel size 3, Conv2 with kernel size 5) to capture temporal/spatial dependencies, followed by a dense layer with ReLU activation and dropout for dimensionality reduction and regularization. The dense output is expanded and passed to a MultiHeadAttention layer to re-weight features, and a residual connection adds this attention output back to the original features, followed by layer normalization to ensure stable training and effective information preservation \cite{xiong2020layer}. The encoded modality representations are  
\begin{equation} \label{eq:h_function}
\mathbf{h}^{(m)} = f_m\big(\mathbf{x}^{(m)}\big)
\end{equation}

where \(\mathbf{h}^{(m)}\) represents the high-level feature embedding of the \(m\)-th modality. To integrate complementary information, we first stack the modality-specific embeddings along a new axis, resulting in a tensor of shape $(M, d')$, where $M$ is the number of modalities and $d'
$ is the embedding dimension. A cross-modal attention mechanism is then applied by computing dependencies between all modalities to refine this representation which enhances inter-modal interactions while suppressing noise from less informative modalities.

\subsubsection{Gating Network}
Following attention refinement, an adaptive gating network assigns importance scores to each modality to ensure that more informative modalities contribute more significantly to the final representation \cite{zhang2024multipath}. From Eq.~\eqref{eq:h_function}, the fused multimodal representation is computed as  
\begin{equation} \label{eq:h_new_function}
\mathbf{h} = \sum_{m=1}^{M} w_m\, \mathbf{h}^{(m)},
\end{equation}
where the learnable weight \(w_m\) for the \(m\)-th modality is defined as  
\[
w_m = \frac{\exp\!\big(\mathbf{W}\, \mathbf{h}^{(m)}\big)}{\sum_{j=1}^{M} \exp\!\big(\mathbf{W}\, \mathbf{h}^{(a)}\big)}
\]
Here, \(\mathbf{W}\) is a learnable parameter matrix that transforms each modality’s embedding \(\mathbf{h}^{(m)}\) into a scalar score, $ \mathbf{h}^{(a)} $ represents the embeddings from all modalities, iterating over $ a \in \{1,2,\dots,M\} $ and \(w_m\) is the normalized weight (via softmax) representing the relative importance of the \(m\)-th modality. This mechanism effectively tackles the challenge of \textbf{heterogeneous feature distributions}. The final representation \(\mathbf{h}\) is then passed to the hyperbolic projection stage.

\subsection{Adaptive Hyperbolic Projection}
The unified multimodal representation \(\mathbf{h}\) is now embedded into the \textbf{Poincaré ball} \(\mathbb{B}^d\), a hyperbolic space that provides improved geometric separation for classification \cite{ma2022adaptive}. Given the fused representation \(\mathbf{h} \in \mathbb{R}^d\), the projection into hyperbolic space is performed as
\[
\mathbf{y} = \tanh\!\big(\alpha\, \|\mathbf{h}\|\big) \cdot \frac{\mathbf{h}}{\|\mathbf{h}\|}
\]
where \(\alpha > 0\) is a trainable curvature parameter and y is the hyperbolic embedding of the multimodal representation \(\mathbf{h}\). This mapping ensures that the output \(\mathbf{y}\) lies within the Poincaré ball \(\mathbb{B}^d\). Unlike fixed-curvature approaches, our model optimizes \(\alpha\) dynamically during training, allowing it to adapt to the data distribution and curvature of the embedding space. Mapping the fused multimodal representation into a shared hyperbolic space ensures that inter-modal relationships are effectively preserved \cite{guo2021multi}.

\subsubsection{Distance Computation in Hyperbolic Space}
Once projected, embeddings are compared using the \textbf{Poincaré distance} \cite{nickel2017poincare}, which measures similarity in hyperbolic space and replaces the standard Euclidean metric used in traditional prototypical networks. The distance between two embeddings \(\mathbf{y}_1,\mathbf{y}_2 \in \mathbb{B}^d\) is computed as:
\begin{equation}
    d_{\mathbb{B}}(\mathbf{y}_1, \mathbf{y}_2) = \operatorname{acosh}\!\left( 1 + \frac{2\, \|\mathbf{y}_1 - \mathbf{y}_2\|^2}{\big(1 - \|\mathbf{y}_1\|^2\big)\big(1 - \|\mathbf{y}_2\|^2\big)} \right)
\end{equation}
Here, \(\operatorname{acosh}(\cdot)\) is the inverse hyperbolic cosine function. This formulation ensures that distances grow exponentially as points move toward the boundary of \(\mathbb{B}^d\), enhancing class separability.

\subsubsection{Residual Hyperbolic Block}
A Residual Hyperbolic Block (RHB) refines the learned embedding by combining the original embedding with a transformed version of itself. Starting with an initial hyperbolic embedding \(\mathbf{y}\), a trainable function \(f(\mathbf{y})\) (e.g., a fully connected layer) is applied to capture high level features. The block then adds the original embedding back to this transformed output, resulting in
\[
\mathbf{y}' = f(\mathbf{y}) + \mathbf{y}
\]
Since we operate in the Poincaré ball \(\mathbb{B}^d\) (a hyperbolic space), the output \(\mathbf{y}'\) is typically re-projected into \(\mathbb{B}^d\) to maintain stability. This residual enables the model to learn richer and more expressive embeddings while ensuring the geometric constraints of hyperbolic space are preserved.

\subsection{Prototype Computation in Hyperbolic Space}
Once both support and query samples have been embedded into the hyperbolic space, prototype-based classification is performed. For a given class \(c\) (with \(c\in\{0,1\}\)), first compute the unweighted prototype:
\begin{equation} \label{eq:proto_un}
\bar{\mathbf{p}}_c = \frac{1}{N_c} \sum_{i=1}^{N_c} \mathbf{y}_i,
\end{equation}
where \(\mathbf{y}_i\) is the hyperbolic embedding of the \(i\)-th support sample and \(N_c\) is the number of support samples in class \(c\). Then, the class prototype \(\mathbf{p}_c\) is computed as a weighted mean in the hyperbolic space from Eq.~\eqref{eq:proto_un} : 
\begin{equation}
    \mathbf{p}_c = \sum_{i} w_i\, \mathbf{y}_i, \quad w_i = \frac{\exp\!\Big(-d_{\mathbb{B}}(\mathbf{y}_i, \bar{\mathbf{p}}_c)\Big)}{\sum_{j} \exp\!\Big(-d_{\mathbb{B}}(\mathbf{y}_j, \bar{\mathbf{p}}_c)\Big)}
\end{equation}
The weights \(w_i\) (obtained via a softmax over negative distances) ensure that points closer to the prototype contribute more. This formulation prevents prototype distortion due to outliers \& maintains class consistency in  hyperbolic space.

\subsection{Loss Functions and Optimization}
To enhance class separability in hyperbolic space, we employ a combination of hyperbolic prototypical loss and angular margin loss, optimizing decision boundaries for few-shot learning.

\subsubsection{Hyperbolic Prototypical Loss}
Given a query embedding \(\mathbf{y}_q\) (representing a query sample in hyperbolic space) and class prototypes \(\mathbf{p}_c\) (the representative embedding for class \(c\)), classification is performed by measuring the Poincaré distance \(d_{\mathbb{B}}(\mathbf{y}_q, \mathbf{p}_c)\) (with \(d_{\mathbb{B}}\) denoting the distance in the Poincaré ball) between the query and each prototype:
\begin{small}
\begin{equation} \label{eq:proto_emb}
d_{\mathbb{B}}(\mathbf{y}_q, \mathbf{p}_c) = \operatorname{acosh}\!\left( 1 + \frac{2\, \|\mathbf{y}_q - \mathbf{p}_c\|^2}{\big(1 - \|\mathbf{y}_q\|^2\big)\big(1 - \|\mathbf{p}_c\|^2\big)} \right)    
\end{equation}
\end{small}
The probability of a query belonging to class \(c\) is then obtained via a softmax over negative distances.
\begin{equation}
p(y_q = c \mid \mathbf{y}_q) = \frac{\exp\!\left(-d_{\mathbb{B}}(\mathbf{y}_q, \mathbf{p}_c)\right)}{\sum_{c'\in\{0,1\}} \exp\!\left(-d_{\mathbb{B}}(\mathbf{y}_q, \mathbf{p}_{c'})\right)}
\end{equation}

The prototypical loss is defined as:
\begin{equation} 
\mathcal{L}_{\text{proto}} = -\mathbb{E} \left[ \log p(y_q = c \mid \mathbf{y}_q) \right],
\label{eq:pr}
\end{equation}

which ensures that query embeddings remain close to their corresponding class prototypes while maximizing inter-class distances \cite{ghadimi2021hyperbolic}.

\subsubsection{Angular Margin Loss}
To further enhance class discrimination, an angular margin loss enforces a margin \(\gamma\) between the cosine similarity of a query with its correct prototype and the most similar incorrect prototype:
\small
\begin{equation} \label{eq:ang}
\mathcal{L}_{\text{angular}} = \mathbb{E} \left[
\max \left( 0, \cos(\theta_{q, p_c}) + \gamma 
- \max_{c' \neq c} \cos(\theta_{q, p_{c'}}) \right)
\right]
\end{equation}
\normalsize

Here, \(c'\) indexes all classes other than the correct class \(c\), and \(\theta_{q, p_c}\) represents the angular distance between the query embedding \(\mathbf{y}_q\) and the correct prototype \(p_c\). The term \(\max_{c' \neq c} \cos(\theta_{q, p_{c'}})\) selects the most similar incorrect prototype, enforcing an angular margin \(\gamma\) between classes. We adopt this loss from ArcFace \cite{deng2019arcface} by incorporating an additive margin \(\gamma\) to increase inter-class separation while preserving intra-class compactness. The total loss is a weighted combination of the hyperbolic prototypical loss and the angular margin loss from Eqs.~\eqref{eq:pr} and~\eqref{eq:ang}:
\begin{equation}
\mathcal{L}_{\text{total}} = \mathcal{L}_{\text{proto}} + \lambda\, \mathcal{L}_{\text{angular}},
\end{equation}
where \(\lambda\) is a weighting parameter that controls the angular margin loss. This combined loss encourages structured hyperbolic embeddings with better class separability.

\section{Experiments}
\subsection{Datasets}
\textbf{Multi-Modal Anxiety Dataset (M2AD)}:
The M2AD dataset was collected at a \textit{anonymous} institute in a low-income Asian country, with ethical approval obtained from the institutional review board. In the study, student participants were invited to a laboratory setting to perform an anxiety-inducing speech task in front of a smartphone camera. Public speaking was chosen as the stressor based on the Trier Social Stress Test (TSST), a widely used protocol for eliciting psychological stress~\cite{kirschbaum1993trier}. Participants were informed that they would speak for two minutes and that their recordings would be evaluated by experts.

On the day of the study, each participant wore Shimmer sensors on their fingers to collect PPG \& EDA signals. Simultaneously, a smartphone (Redmi 9A), mounted on a tripod, recorded video of the participant (see Figure~\ref{fig:study}). A research assistant randomly selected a topic from a psychiatrist-approved list (see Appendix) \& instructed the participant to speak immediately, without any preparation time. The speech duration was fixed at two minutes.

Following the task, participants completed the anxiety subscale of the Depression, Anxiety, and Stress Scale (DASS-21) \cite{lovibond1995manual}, which includes seven items rated on a 4-point Likert scale (0 = ``Not at all" to 3 = ``Very much"). Responses were summed to compute an overall anxiety score. Participants scoring below 10 were labeled as ``non-anxious," while those scoring 10 or higher were as ``anxious," following validated DASS-21 thresholds.

After the study, PPG, EDA, and the recorded video of the Speech task were extracted for further analysis. A total of 108 (60 anxious and 48 non-anxious) participants took part in our study, including 81 males and 27 females. The average age of the participants was 19.73 ($\pm$1.34). As a token of appreciation for their time, they received a refreshment box upon completing the study.

\noindent\textbf{Social Anxiety Dataset (SAD)}: 
To evaluate the robustness HCFSLN, we tested it on Social Anxiety Dataset (SAD)~\cite{sahu2024beyond,sahu2024unveiling,sahu2024wearable}, which includes all modalities present in M2AD. In the SAD study, participants performed a 2.5-minute anxiety provoking speech activity in front of an audience while their ECG, PPG, and EDA signals were passively recorded. Additionally, the speech activity was captured using an audiovisual camera. We used data from 92 participants (41 anxious and 51 non-anxious) to benchmark our framework for anxiety detection. 

\subsection{Implementation Details} \label{section: data preprocessing}
\textbf{Data preprocessing:}
Though the speech activity in \textbf{M2AD} was of 120 seconds, some instances had slight variations of approximately ($\pm$) 5 seconds in the data. To address this, we standardized the data length to 120 seconds. So, if any data modality exceeded 120 seconds, we trimmed it to the first 120 seconds, and if the data length was shorter than 120 seconds, we padded it with zeros to make it up to 120 seconds. For feature extraction in audio and video modalities, we used a non-overlapping 1-second window where the feature extraction was done for each window. For the video modality, we used OpenFace\footnote{https://github.com/TadasBaltrusaitis/OpenFace} to extract 711 features related to eye and head position, face landmarks, and facial action units \cite{baltruvsaitis2016openface}. In audio, we followed the methodology of Sahu et al. and Mohammadi et al. to extract 113 audio features using the Librosa Python package \footnote{https://librosa.org/doc/latest/index.html} \cite{mohammadi2017robust,sahu2024unveiling}. For the PPG signal, we first cleaned the noisy data using the NeuroKit Python package \footnote{https://github.com/neuropsychology/NeuroKit}, then downsampled the cleaned signal from 512Hz to 1Hz to match the shape of the audio and video inputs \cite{makowski2021neurokit2}. Similarly, the EDA signal was first cleaned using a Butterworth filter from NeuroKit and decomposed into phasic and tonic components, which were also downsampled from 512Hz to 1Hz to match the audio and video input shapes. \textbf{SAD} dataset was processed in a similar manner as mentioned above. The extracted features are concatenated into a shape \( (N, L, D) \), where \( N \) represents the number of participants, \( L \) is the fixed sequence length of 120, and \( D \) is the total feature dimension across modalities. This structure preserves temporal dependencies while enabling efficient sequence-based learning \cite{menon2021multimodal}.

\subsection{Baseline Models}

We benchmarked HCFSLN against \textbf{six} recent SOTA FSL networks: Few-shot Sequence Learning with Transformers (TAM)~\cite{logeswaran2020few}—adapted with LSTM encoders for temporal sequence data; Uniform Hyperspherical Structure-preserving Embeddings (noHub)~\cite{trosten2023hubs}; Hyperbolic Few-Shot Contrastive Learning (HCL)~\cite{choi2024differentiating}—using LSTM encoders; Efficient Spatiotemporal Few-Shot Learning in Hyperbolic Space (ES-HMS)~\cite{wei2024hms}—implemented with LSTM modality encoders and hyperbolic prototype fusion; EmoTracer~\cite{kuang2024emotracer}—LSTM-based fusion; and Contrastive Prototypical Learning (CPL-Net)~\cite{wang2025multi}—using transformer-based temporal attention for sequential data. These SOTA baselines were selected to represent the major FSL approaches using Euclidean, hyperspherical, and hyperbolic embeddings. We trained and evaluated each method under both one-shot and five-shot settings to assess performance across varying sample sizes.

\begin{table*}[!t]
\centering

\small  
\begin{adjustbox}{width=\textwidth}
\begin{tabular}{l c c c c c c c c c}
\toprule
\textbf{Model} & \textbf{Shots} & \multicolumn{4}{c}{\textbf{SAD}} & \multicolumn{4}{c}{\textbf{M2AD}} \\
\cmidrule(lr){3-6} \cmidrule(lr){7-10}
& & \textbf{Single} & \textbf{Dual} & \textbf{Triple} & \textbf{All} & \textbf{Single} & \textbf{Dual} & \textbf{Triple} & \textbf{All} \\
\midrule
TAM & 1 & A(0.63, 0.05) & E-P(0.66, 0.03) & E-P-V(0.63, 0.07) & (0.61, 0.03) & V(0.69, 0.06) & P-V(0.74, 0.04) & E-P-V(0.69, 0.10) & (0.66, 0.07) \\
& 5 & P(0.66, 0.03) & E-V(0.66, 0.06) & A-E-P(0.67, 0.04) & (0.64, 0.06) & V(0.66, 0.14) & P-V(0.67, 0.13) & E-P-V(0.71, 0.11) & (0.63, 0.08) \\
\addlinespace
NoHub & 1 & A(0.71, 0.03) & A-E(0.67, 0.04) & A-E-P(0.65, 0.08) & (0.55, 0.13) & V(0.66, 0.10) & A-E(0.68, 0.07) & E-P-V(0.66, 0.11) & (0.66, 0.05) \\
& 5 & A(0.70, 0.10) & A-E(0.65, 0.06) & E-P-V(0.55, 0.13) & (0.52, 0.04) & V(0.70, 0.11) & A-E(0.66, 0.13) & A-E-P(0.66, 0.12) & (0.65, 0.08) \\
\addlinespace
HCL & 1 & V(0.73, 0.06) & E-P(0.68, 0.07) & A-E-P(0.71, 0.08) & (0.66, 0.09) & V(0.65, 0.06) & P-V(0.68, 0.07) & A-P-V(0.67, 0.13) & (0.57, 0.05) \\
& 5 & P(0.71, 0.05) & A-P(0.67, 0.07) & E-P-V(0.63, 0.06) & (0.60, 0.06) & E(0.69, 0.11) & E-P(0.68, 0.09) & A-E-P(0.65, 0.05) & (0.58, 0.07) \\
\addlinespace
ES-HMS & 1 & P(0.68, 0.00) & E-P(0.72, 0.05) & E-P-V(0.65, 0.06) & (0.66, 0.06) & V(0.64, 0.05) & P-V(0.70, 0.06) & A-E-P(0.67, 0.09) & (0.63, 0.07) \\
& 5 & P(0.68, 0.04) & E-P(0.66, 0.13) & A-E-V(0.63, 0.08) & (0.63, 0.05) & E(0.69, 0.06) & E-P(0.66, 0.07) & E-P-V(0.64, 0.07) & (0.57, 0.06) \\
\addlinespace
EmoTracer & 1 & P(0.67, 0.06) & E-P(0.66, 0.04) & E-P-V(0.62, 0.10) & (0.67, 0.10) & E(0.65, 0.06) & E-P(0.62, 0.12) & A-E-P(0.60, 0.09) & (0.60, 0.12) \\
& 5 & P(0.71, 0.10) & E-P(0.66, 0.10) & A-P-V(0.60, 0.06) & (0.58, 0.10) & P(0.65, 0.14) & A-P(0.61, 0.06) & A-E-V(0.63, 0.05) & (0.65, 0.13) \\
\addlinespace
CPL-Net & 1 & P(0.70, 0.04) & P-V(0.70, 0.04) & E-P-V(0.72, 0.08) & (0.63, 0.08) & E(0.59, 0.08) & A-P(0.62, 0.12) & A-P-V(0.62, 0.11) & (0.56, 0.05) \\
& 5 & V(0.66, 0.08) & A-P(0.65, 0.03) & A-P-V(0.67, 0.06) & (0.62, 0.04) & E(0.60, 0.05) & A-P(0.66, 0.05) & A-P-V(0.67, 0.06) & (0.62, 0.06) \\
\addlinespace
\textbf{HCFSLN (Ours)} & 1 & \textbf{A(0.88, 0.02)} &\textbf{ A-E(0.85, 0.07)} & \textbf{A-P-V(0.85, 0.06)} & \textbf{(0.79, 0.07)} & \textbf{A(0.79, 0.07) }& A-\textbf{P(0.82, 0.08) }& \textbf{A-E-V(0.83, 0.02) }& \textbf{(0.75, 0.05) }\\
& 5 & \textbf{A(0.82, 0.07)} & \textbf{A-E(0.82, 0.04)} & \textbf{A-P-V(0.86, 0.05)} & \textbf{(0.80, 0.05)} & \textbf{A(0.78, 0.06) }&\textbf{ E-V(0.81, 0.03) }&\textbf{ A-E-V(0.79, 0.09)} &\textbf{ (0.78, 0.03)} \\
\bottomrule
\end{tabular}
\end{adjustbox}
\caption{Performance comparison of models across modalities (single, dual, triple, and all) and shot settings (1-shot and 5-shot). Modalities—Audio, Video, PPG, and EDA—are abbreviated as A, V, P, and E, respectively. For each model, only the best-performing modality combination(s) are shown under 1 and 5-shot settings.} \label{tab:performance}
\end{table*}

\paragraph{Few-Shot Training Configuration:} StandardScaler normalization, sequence padding, and stratified train-test splitting (\textit{test\_size} = 0.2) are applied across all models. Training is conducted using the Adam optimizer (\textit{learning rate} = $1e^{-3}$, 50 epochs), with each episode selecting \(K\) support samples (\(K \in \{1,5\}\)) per class (anixous, non-anxious) . In the proposed FSL framework, a residual hyperbolic block with an adaptive projection layer (trainable curvature initialized as \(\alpha = 1.0\)) refines representations. Optimization combines hyperbolic prototypical loss and angular margin loss (\(\gamma\) = 0.2, \(\lambda\) = 1.0) to enhance class separability. Each experiment is repeated 5 times, reporting average accuracy with standard deviations \cite{fu2022worst} for model stability and robustness.   \textit{All experiments were conducted on a system equipped with an NVIDIA RTX 4070 Ti 12GB GPU, 64 GB RAM, Win 11 and an Intel i9 12th Gen CPU.}

\begin{figure}[!t]
    \centering
    \captionsetup{font=small} 

    \begin{subfigure}[b]{0.32\linewidth}
        \includegraphics[width=\linewidth]{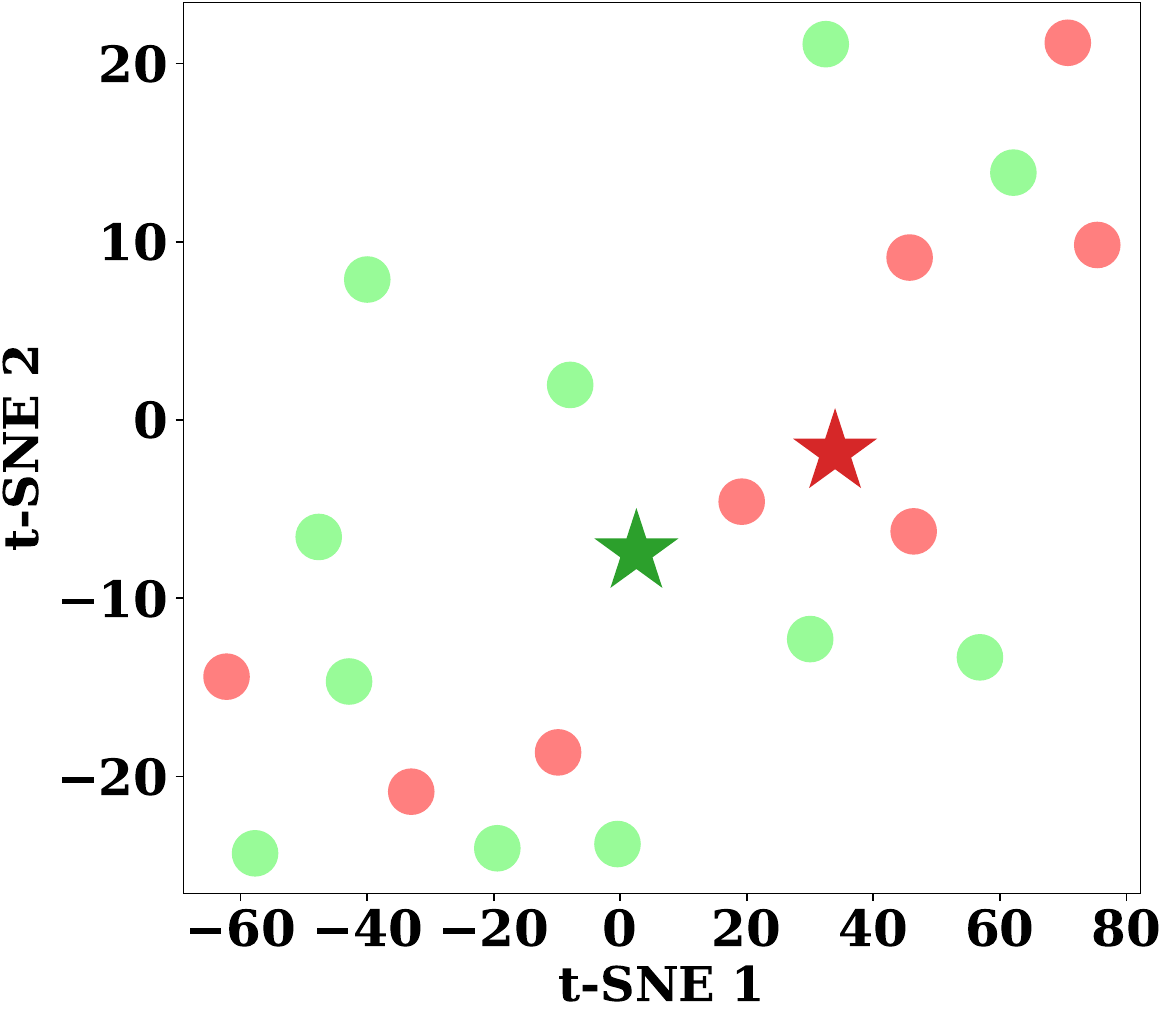}
        \caption{CPL-Net}
    \end{subfigure}
    \hfill
    \begin{subfigure}[b]{0.32\linewidth}
        \includegraphics[width=\linewidth]{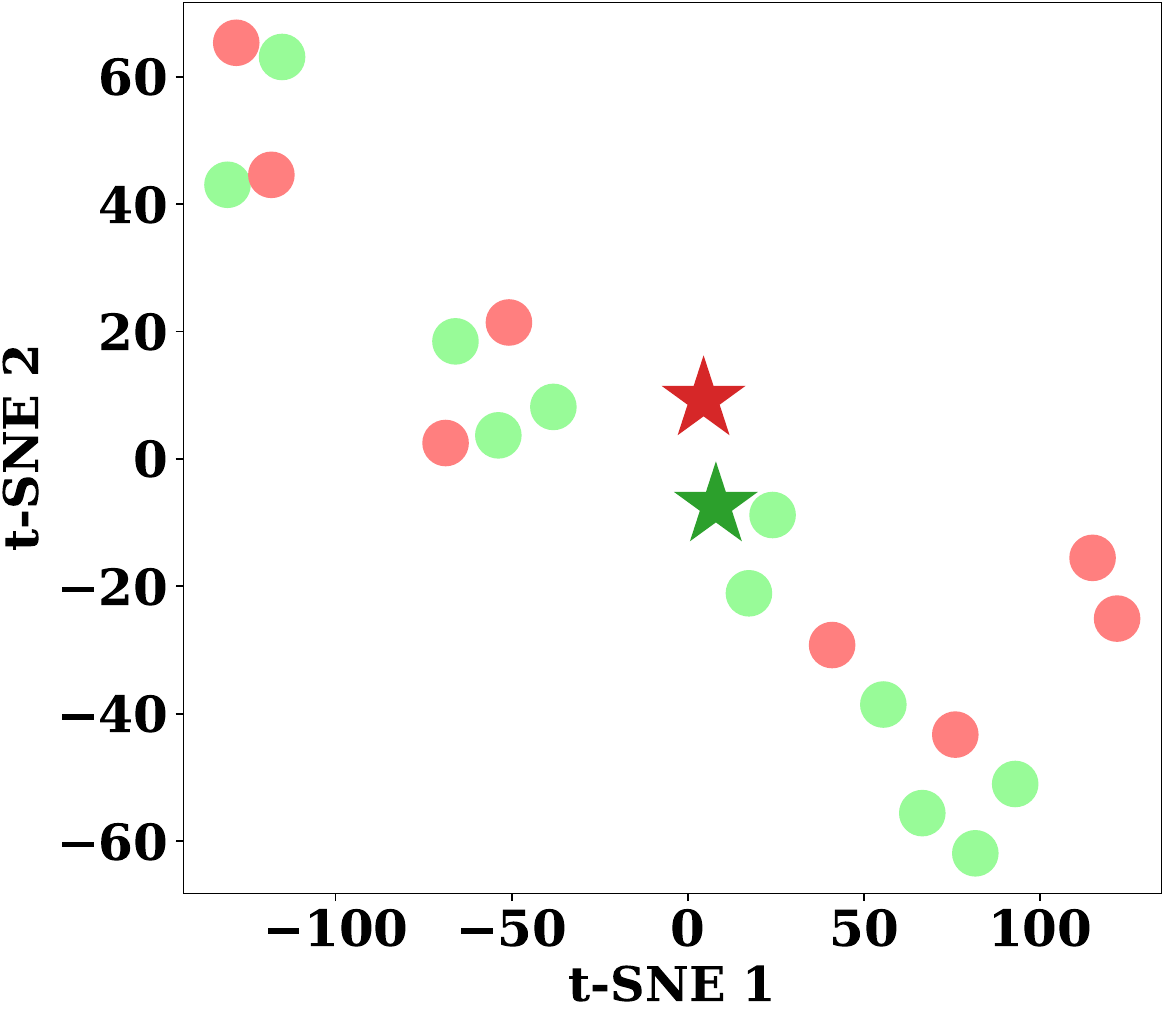}
        \caption{EmoTracer}
    \end{subfigure}
    \hfill
    \begin{subfigure}[b]{0.32\linewidth}
        \includegraphics[width=\linewidth]{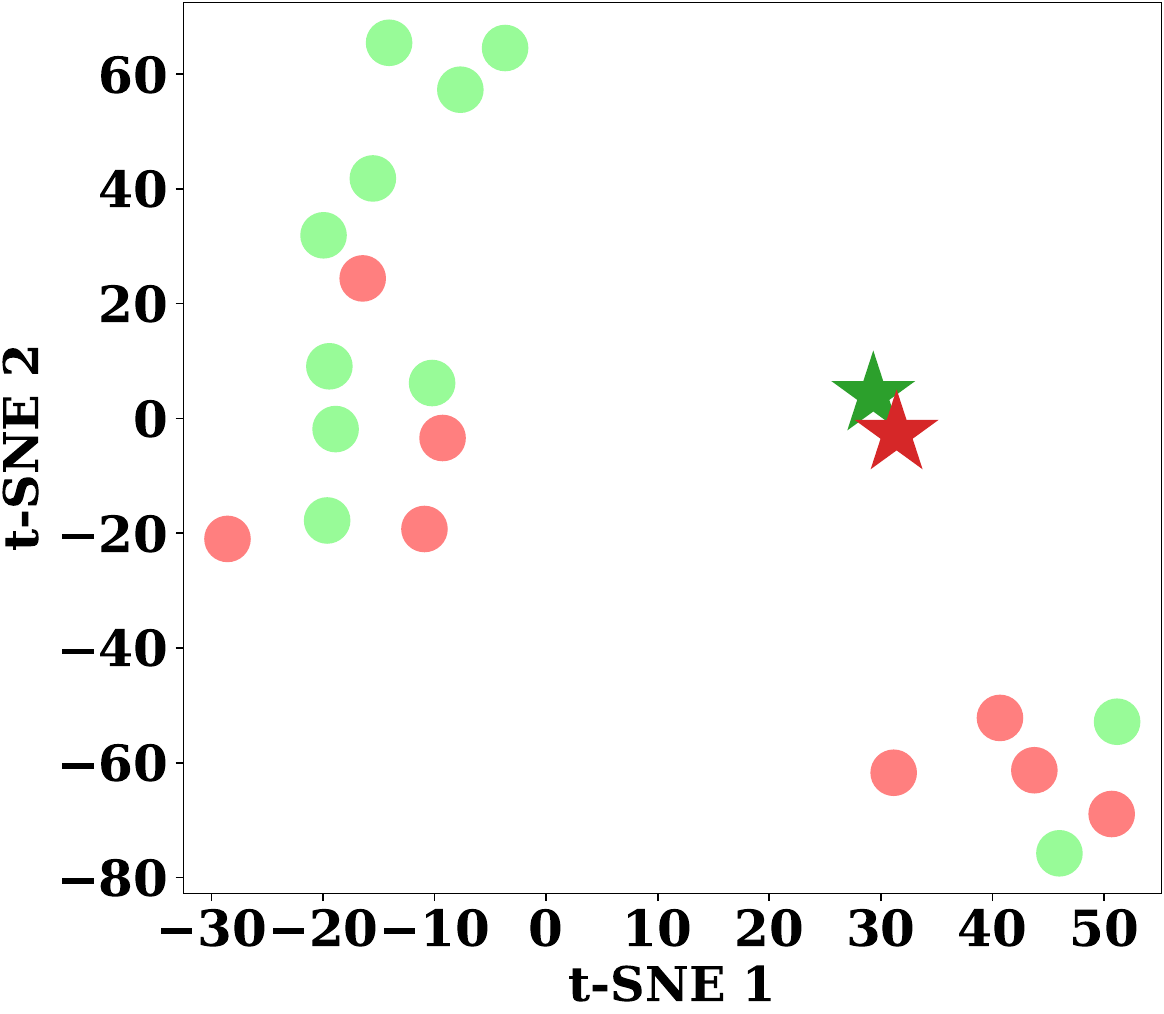}
        \caption{ES-HMS}
    \end{subfigure}

    \vspace{0.3cm} 

    \begin{subfigure}[b]{0.32\linewidth}
        \includegraphics[width=\linewidth]{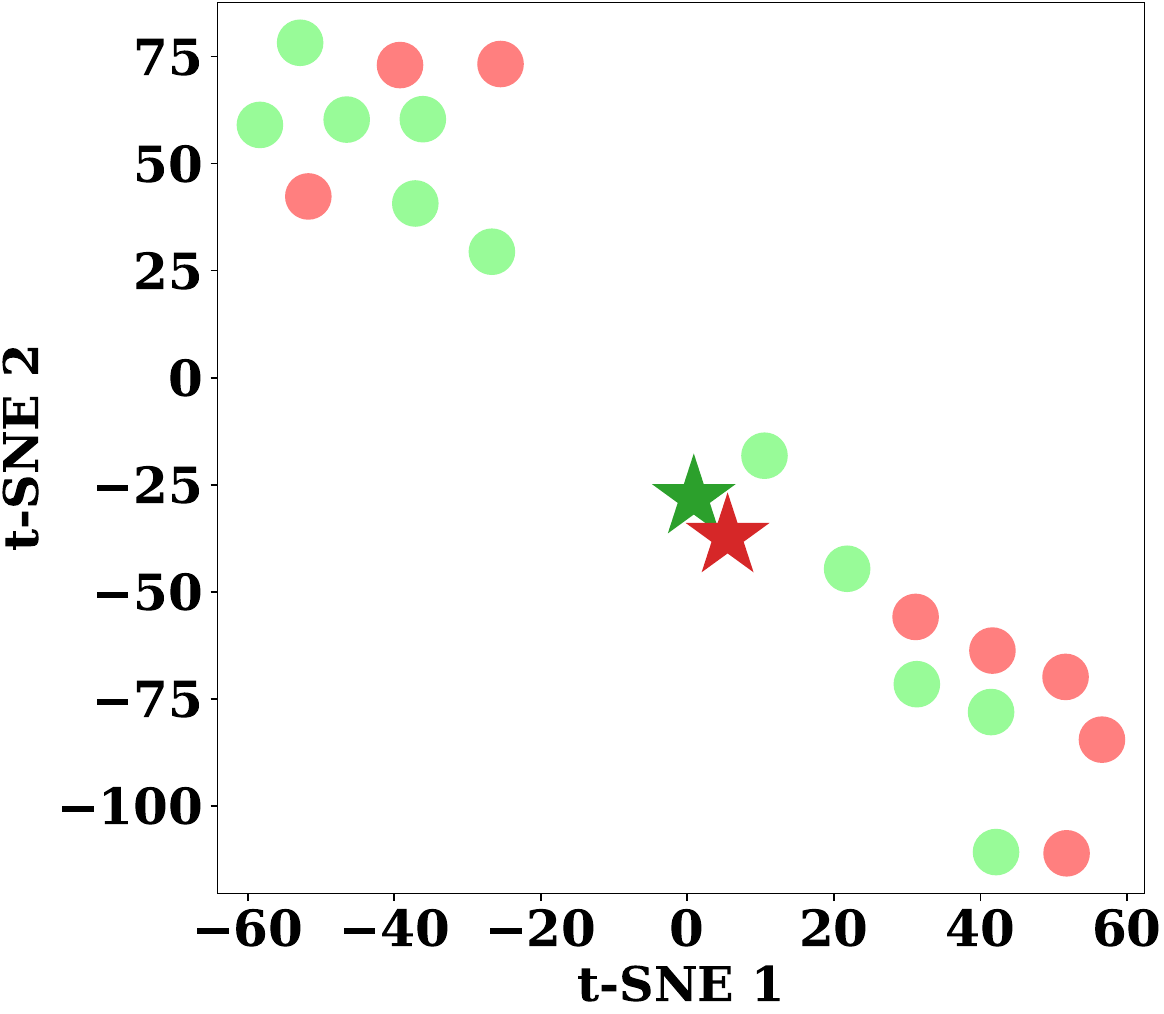}
        \caption{HCL}
    \end{subfigure}
    \hfill
    \begin{subfigure}[b]{0.32\linewidth}
        \includegraphics[width=\linewidth]{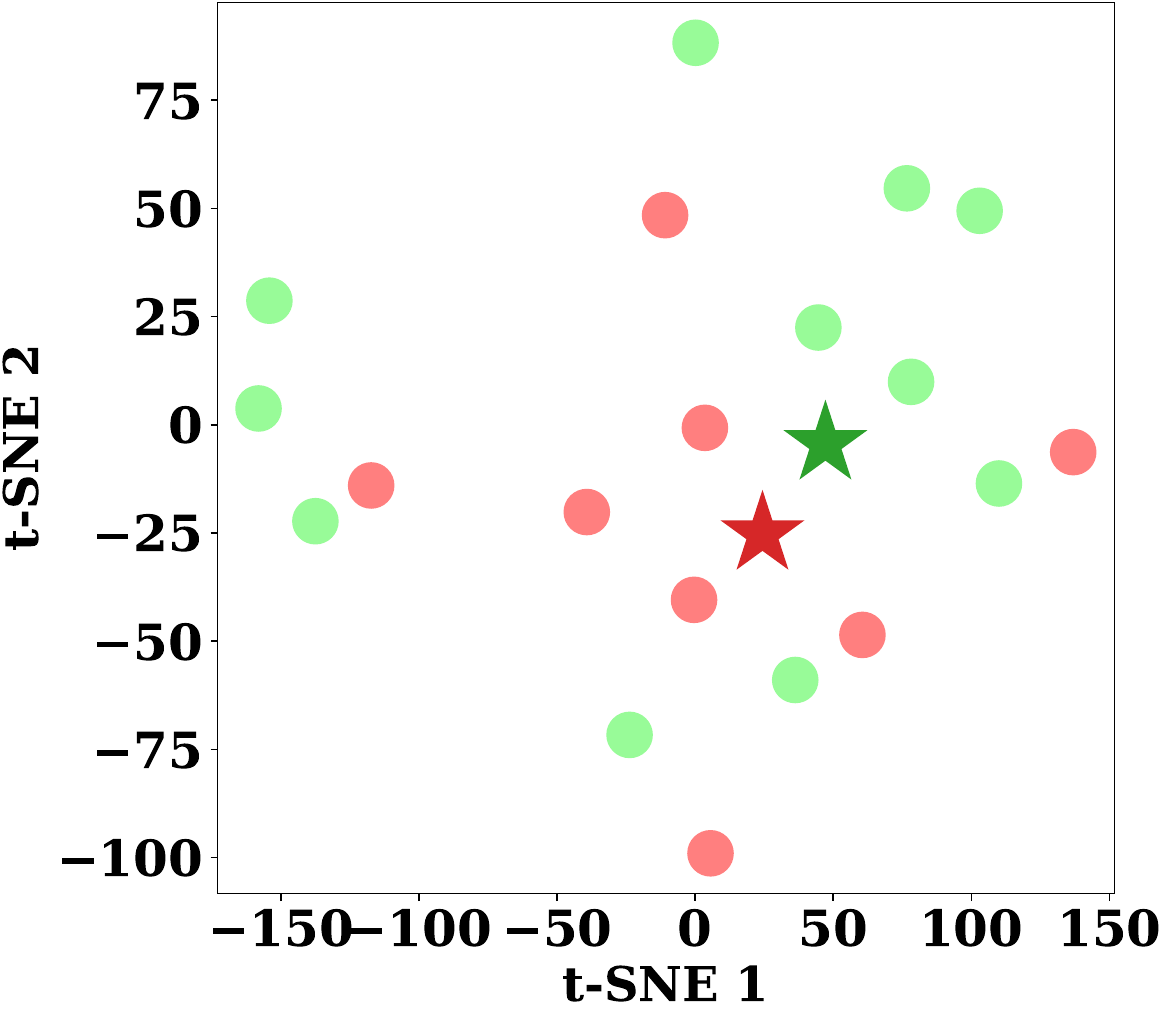}
        \caption{NoHub}
    \end{subfigure}
    \hfill
    \begin{subfigure}[b]{0.32\linewidth}
        \includegraphics[width=\linewidth]{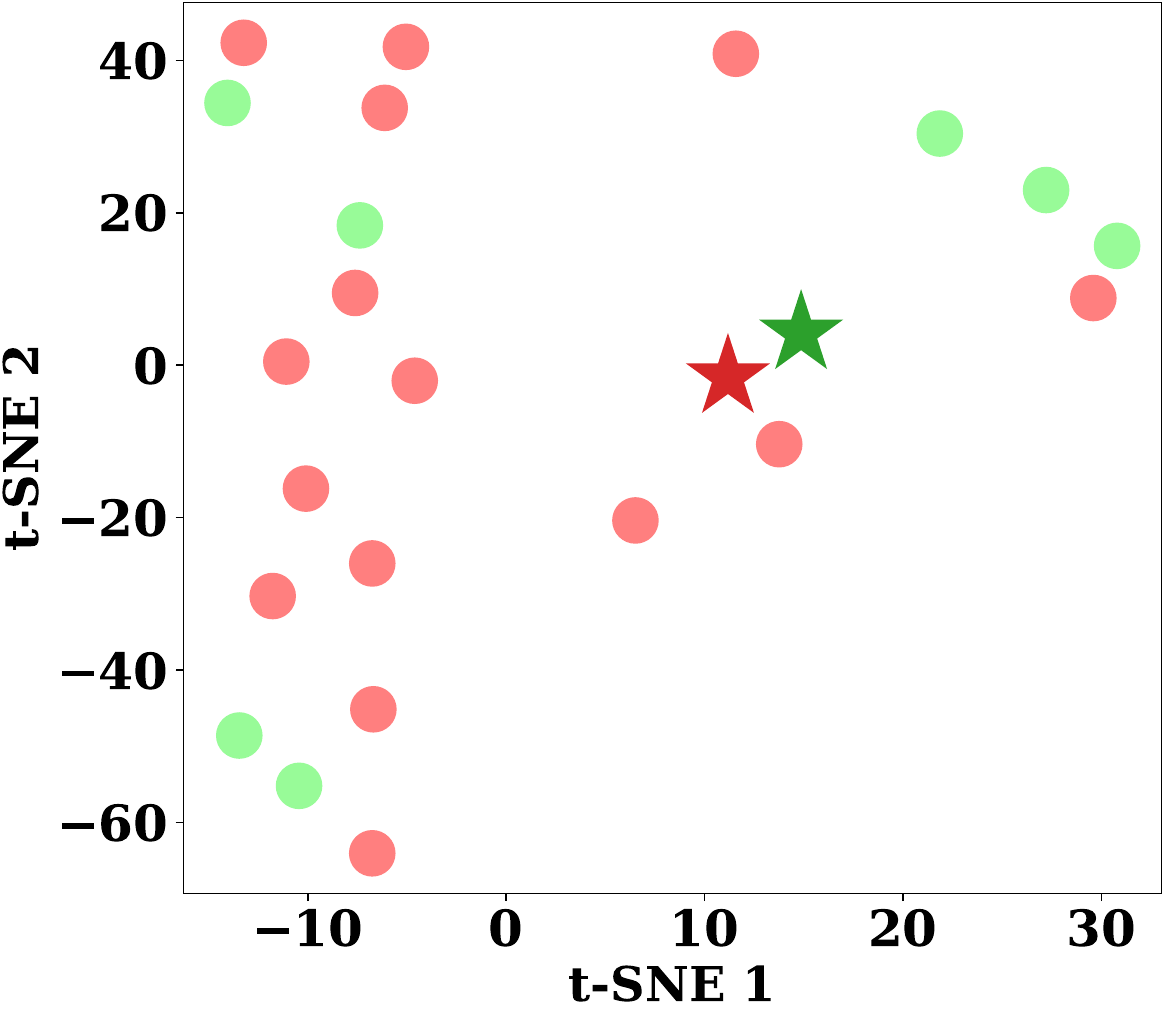}
        \caption{TAM}
    \end{subfigure}

    \vspace{0.3cm} 

    \begin{subfigure}[b]{0.32\linewidth}
        \centering
        \includegraphics[width=\linewidth]{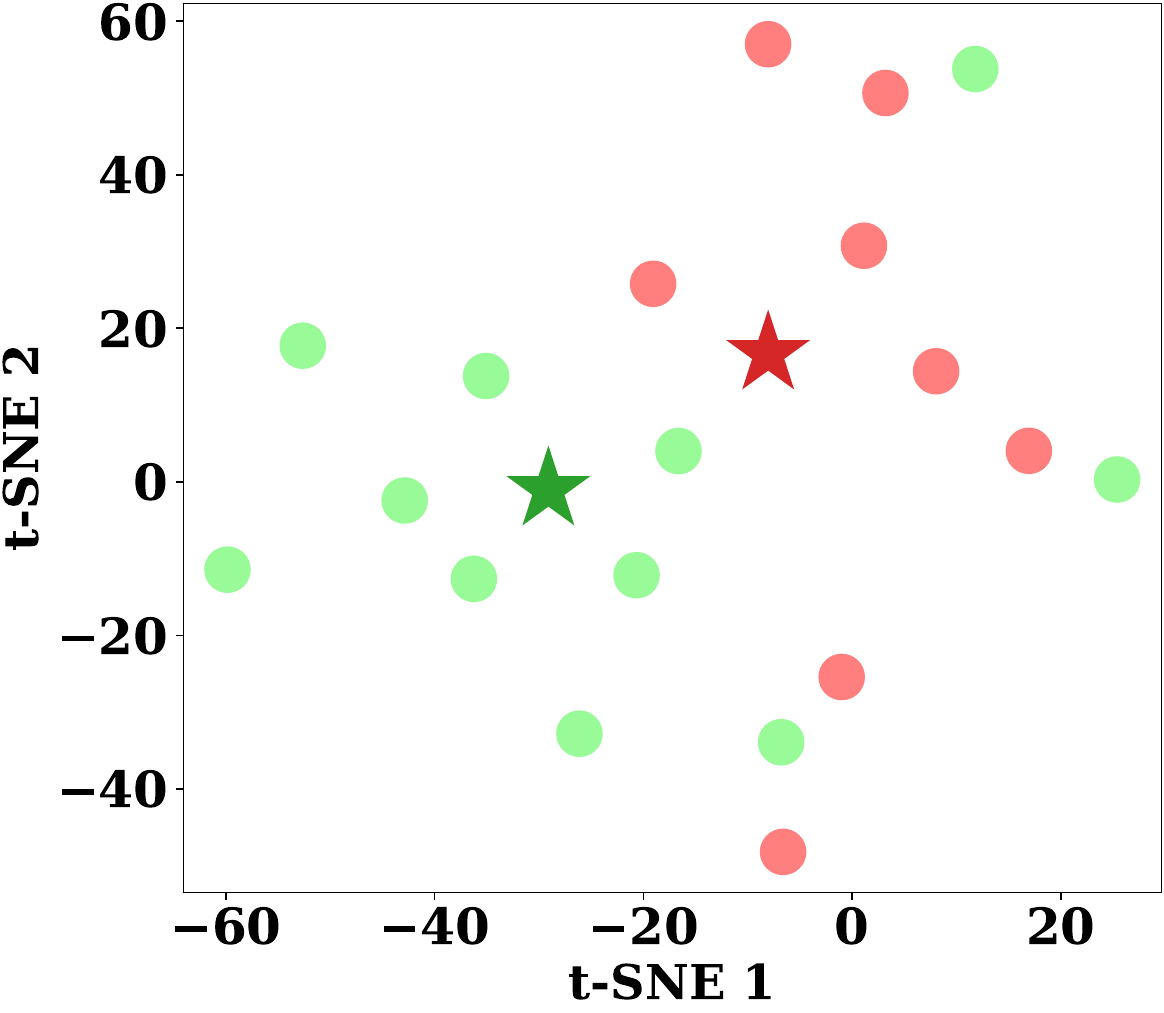}
        \caption{HCFSLN (Ours)}
    \end{subfigure}

    \caption{t-SNE prototype visualizations of models.}
    \label{fig:tsne_visualizations}
\end{figure}

\begin{figure*}[t]
    \centering
    \begin{subfigure}[t]{0.32\linewidth}
        \centering
        \includegraphics[width=\linewidth]{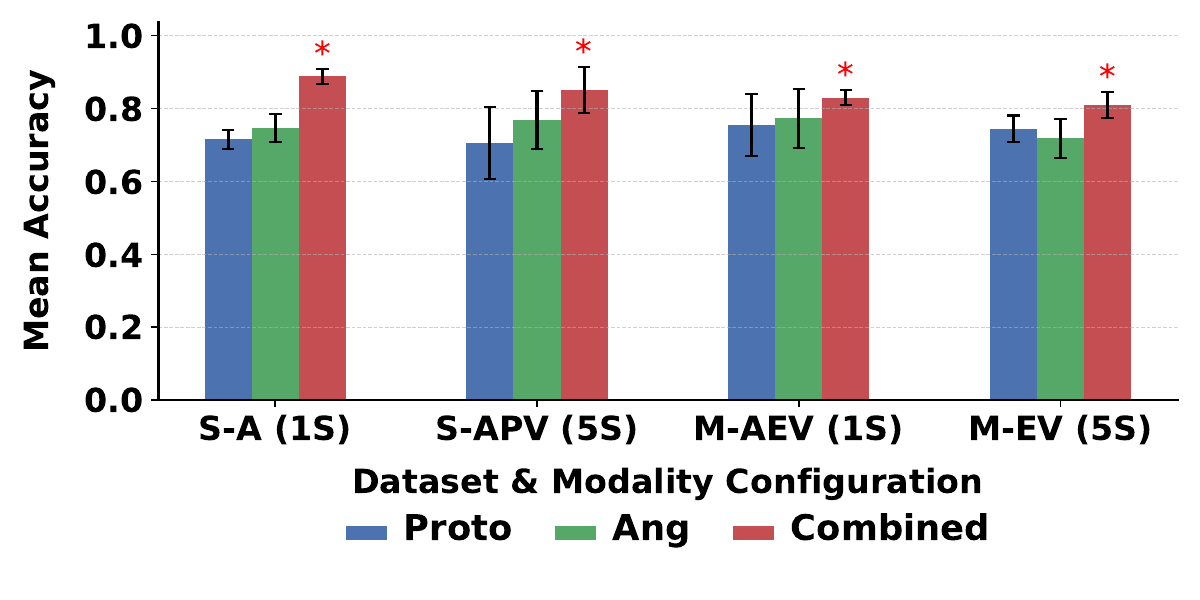}
        \caption{Impact of loss type combinations.}
        \label{type}
    \end{subfigure}
    \hfill
    \begin{subfigure}[t]{0.32\linewidth}
        \centering
        \includegraphics[width=\linewidth]{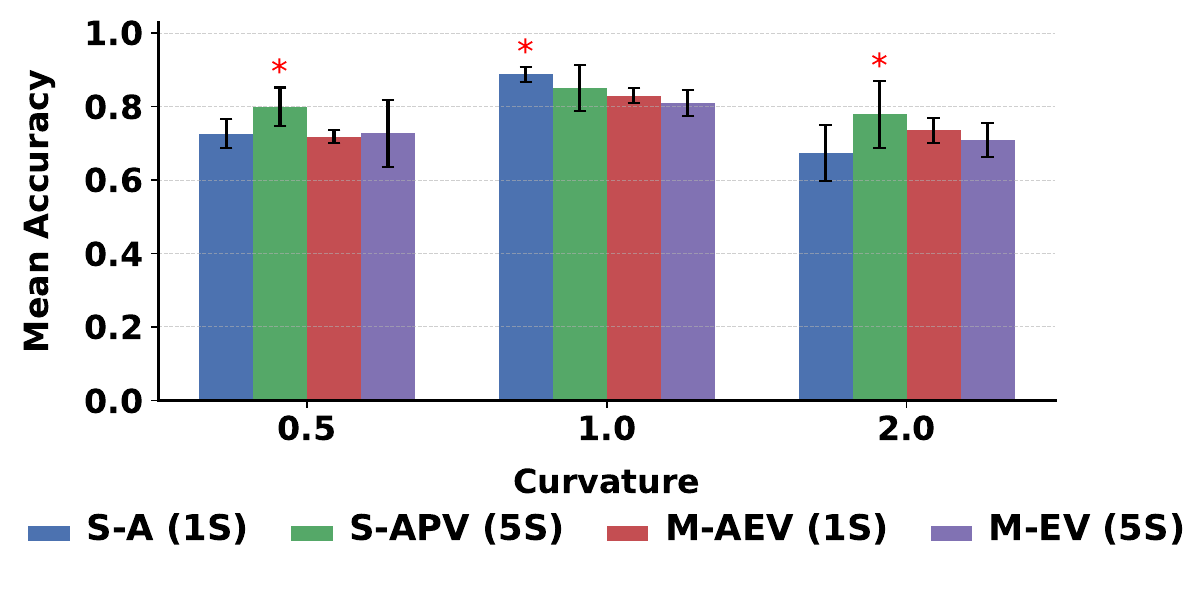}
        \caption{Impact of varying curvature value.}
        \label{fig:curv}
    \end{subfigure}
    \hfill
    \begin{subfigure}[t]{0.32\linewidth}
        \centering
        \includegraphics[width=\linewidth]{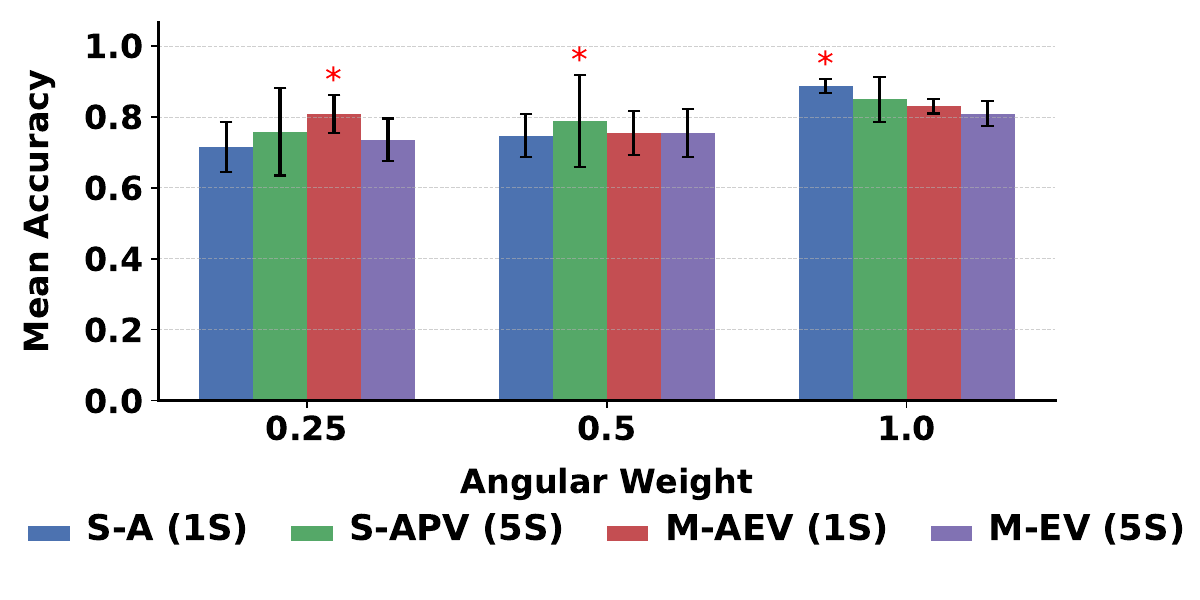}
        \caption{Impact of varying angular weights $\lambda_{\text{ang}}$}
        \label{fig:angular-ablation}
    \end{subfigure}
    \caption{Ablation results showing the effect of loss type, curvature, and angular weight on accuracy.}
    \label{fig:ablation_study}
\end{figure*}
\subsection{Results} 

Table~\ref{tab:performance} highlights clear, quantifiable trends across models, modalities, and shot settings on the SAD and M2AD datasets, presenting the best performing modality combinations. HCFSLN consistently outperforms all baselines across datasets, modalities, and shot settings, proving its effectiveness for few-shot anxiety detection. In single-modality settings under low-shot (1-shot), audio (A) performs best, achieving 88\% accuracy on SAD and 79\% on M2AD, highlighting its strong predictive power as a non-intrusive behavioral modality. For dual modalities, combinations involving audio and physiological signals such as Audio-EDA (A-E) and Audio-PPG (A-P) improve accuracy by approximately 3--5\% over unimodal audio, demonstrating the complementary value of adding intrusive physiological modalities captured via wearable sensors. Tri-modal fusion, combining audio, physiological, and video data (e.g., A-P-V or A-E-V), achieves the highest accuracies---up to 86\% on SAD and 83\% on M2AD---outperforming unimodal and dual setups by 5--7\%, underscoring the advantage of leveraging multiple data sources. Interestingly, increasing from 1-shot to 5-shot settings generally maintains or slightly improves performance, with tri-modal fusion showing the most stable or improved accuracy, while some unimodal and dual-modal results experience minor decreases (around 4--6\%), likely due to data complexity. Also using all modality often reduce accuracy due to addition of extra noise and data complexity. Across all modalities, HCFSLN exhibits significantly lower variance, reflecting more consistent and reliable learning compared to other models. 

\textit{Overall non-intrusive modalities (audio, video) offer a strong baseline, while adding intrusive physiological signals significantly boosts accuracy—especially in multimodal fusion—highlighting the importance of modality choice based on data availability and practical constraints.}

\section{Prototype Visualizations}

Figure~\ref{fig:tsne_visualizations} presents the best prototype t-SNE visualizations for each model, selected from five runs based on highest overall accuracy to ensure a fair comparison. The plots illustrate clusters centered on their prototypes. HCFSLN exhibits particularly compact clusters with centrally located prototypes, reflecting strong discriminative ability. In contrast, models using Euclidean embeddings (EmoTracer, TAM, CPL-Net) display more dispersed and overlapping clusters, while hyperspherical embeddings (NoHub) show moderate clustering but less distinct class separation. Hyperbolic embeddings (ES-HMS, HCL, HCFSLN) effectively capture complex data relationships, as evidenced by their tighter, well-separated clusters, highlighting their superior representational power for few-shot anxiety detection.


\section{Ablation Studies}
We evaluated how loss functions, angular margin weight ($\lambda$), and hyperbolic curvature ($\alpha$) affect HCFSLN accuracy in 1-shot and 5-shot settings for the best configurations. We picked best configuration of HCFSLN in both  SAD and M2AD datasets as SAD-Audio (S-A) (1 shot), SAD Audio-PPG-Video (S-APV) (5 shot), M2AD Audio-EDV-Video (M-AEV) (1  shot), and M2AD EDA-Video (M-EV) (5 shot). We performed Welch’s t-tests \cite{sakai2016two} to evaluate the significance of all three ablation loss types, as well as the effects of hyperbolic curvature and angular weights.

\paragraph{Impact of Loss Type:}

Fig.~\ref{type} compares the effects of Hyperbolic Prototypical Loss (Proto), Angular Loss (Ang), and their combination (Combined) on accuracy across datasets and modalities. The combined loss consistently outperforms individual losses. Specifically, (S-A) (1 shot) combined significantly surpasses both proto and angular losses ($p < 0.001$). For (M-AEV) (1 Shot), combined outperforms proto alone ($p = 0.036$), and in (M-EV) (5 shot), it significantly improves over angular loss ($p = 0.012$). No significant difference was observed in the (S-APV) (5 Shot) setting. Overall, combining hyperbolic and angular margin losses improves robustness and accuracy in few-shot settings.

\paragraph{Impact of Hyperbolic Curvature ($\alpha$):}

We assessed the impact of curvature on HCFSLN by comparing accuracy means at $\alpha = 0.5, 1.0,$ and $2.0$ across datasets and modality configurations. As shown in Fig.~\ref{fig:curv}, for (S-A) (1-shot) accuracy increased by 17.5\% ($p=0.00007$) from ($\alpha$) 0.5 to 1.0, then declined significantly at 2.0 compared to 1.0 ($p=0.0026$). In (S-APV) (5-shot), accuracy peaked at $\alpha = 1.0$ with stable performance and no significant differences across curvatures ($p > 0.18$). For (M-AEV) (1-shot), accuracy improved by 15.5\% from ($\alpha$) 0.5 to 1.0 ($p = 0.001$), with a marginally non-significant change between ($\alpha$) 1.0 and 2.0 ($p = 0.06$). The (M-EV) (5-shot) setting showed moderate, non-significant gains at 1.0 over 0.5 ($p = 0.24$). Overall, $\alpha = 1.0$ consistently achieves the highest and most robust accuracy, underscoring the crucial role of hyperbolic curvature in enhancing representations.

\paragraph{Imapact of Angular Weights:}

In (Fig.~\ref{fig:angular-ablation}) we analyze the effect of angular loss weight ($\lambda$) on accuracy . Significant accuracy improvements were observed when increasing $\lambda$ from 0.25 to 1.0 in all settings, with gains up to 17.2\% (e.g., $p=0.0007$ for (S-A) (1-shot)), confirming the benefit of angular loss. Comparisons between 0.25 and 0.5, and 0.5 and 1.0 showed mixed effect ($p$ ranging 0.02–0.25), indicating a gradual but performance increase as $\lambda$ grows. These results shows that a combined loss with a higher angular weight optimally balances hyperbolic and angular components, enhancing accuracy in few-shot multimodal anxiety detection.






\section{Use Cases and Social Impact}
One in eight people worldwide lives with a mental disorder, with anxiety among the most prevalent \cite{who_report1}. The World Health Organization reports a median of just 13 mental health workers per 100,000 people globally, heavily concentrated in high-income countries with over 40 times more workers than low-income regions. Africa and Southeast Asia, for instance, have only 1.6 and 2.8 workers per 100,000, leading to long waits for support. Economic constraints and GDP priorities often limit investment in these areas. Our research pioneers anxiety detection using wearable sensors (PPG, EDA) and audiovisual data from low-end smartphones, enabling early screening and timely intervention without costly equipment.

\section{Conclusion}

In this work, we introduced HCFSLN for multimodal anxiety detection, addressing the challenge of small mental health datasets. By integrating speech, physiological signals (PPG, EDA), and video features, our framework leverages hyperbolic embeddings, cross-modal attention, and an adaptive gating network to enhance feature separability and generalization. Experiments show HCFSLN outperforms existing few-shot learning models. It can be integrated into wearables and smartphone apps for real-time anxiety assessment, enabling early detection and intervention. This benefits mental health professionals, workplaces, and remote healthcare by providing accessible, proactive anxiety monitoring.

\bibliography{ijcai25}

\end{document}